\documentclass[runningheads]{llncs}

\usepackage{eccv}

\usepackage{eccvabbrv}

\usepackage{graphicx}
\usepackage{booktabs}
\usepackage{multirow}

\usepackage[accsupp]{axessibility}  

\usepackage[pagebackref,breaklinks,colorlinks,citecolor=eccvblue]{hyperref}

\usepackage{orcidlink}

\begin{document}

\title{Elysium: Exploring Object-level Perception in Videos via MLLM} 

\titlerunning{Elysium}

\author{Han Wang \and
Yanjie Wang \and
Yongjie Ye \and
Yuxiang Nie \and
Can Huang
}

\authorrunning{H. Wang et al.}

\institute{Bytedance Inc.
}

\newcommand{\model}[1]{Elysium}
\newcommand{\dataset}[1]{ElysiumTrack-1M}

\maketitle

\begin{abstract}
Multi-modal Large Language Models (MLLMs) have demonstrated their ability to perceive objects in still images, but their application in video-related tasks, such as object tracking, remains understudied. This lack of exploration is primarily due to two key challenges. Firstly, extensive pretraining on large-scale video datasets is required to equip MLLMs with the capability to perceive objects across multiple frames and understand inter-frame relationships. Secondly, processing a large number of frames within the context window of Large Language Models (LLMs) can impose a significant computational burden.
To address the first challenge, we introduce \textbf{\dataset{}}, a large-scale video dataset supported for three tasks: Single Object Tracking (SOT), Referring Single Object Tracking (RSOT), and Video Referring Expression Generation (Video-REG). \dataset{} contains 1.27 million annotated video frames with corresponding object boxes and descriptions. Leveraging this dataset, we conduct training of MLLMs and propose a token-compression model \textbf{T-Selector} to tackle the second challenge. Our proposed approach, \textbf{Elysium}: Exploring Object-level Perception in Videos via MLLM, is an end-to-end trainable MLLM that attempts to conduct object-level tasks in videos without requiring any additional plug-in or expert models. All codes and datasets are released at \url{https://github.com/Hon-Wong/Elysium}.
\keywords{Multi-modal Large Language Models \and Object Tracking \and Referring Single Object Tracking}
\end{abstract}
\section{Introduction}
In recent years, Multi-modal Large Language Models~\cite{blip2,cogvlm,llava,kosmos2,llava15,minigpt,minigpt2} have gained significant attention and proven effective in various multi-modal tasks. These models~\cite{minigpt2,shikra,gpt4roi,cogvlm,qwenvl} have showcased their ability to excel in tasks like Image Grounding~\cite{refcoco,refcocog} and Object Detection~\cite{coco,objects365}, demonstrating their capacity to handle object-level tasks in static images, incorporating an adapter that bridges a visual encoder with a Large Language Model (LLM)~\cite{vicuna2023,llama}.
However, when it comes to video scenes, certain challenges arise due to the complexity of temporal information and the abundance of motion cues, making it challenging for a unified MLLM to effectively address diverse object-level tasks in videos.

Specifically, we classify video tasks into three categories based on the granularity they address: 1) video-level tasks (\textit{e.g.}, VideoQA~\cite{xu2017msrvttqa,yu2019activityqa}  Video Caption~\cite{msrvtt,msvd}), 2) frame-level tasks (\textit{e.g.}, Video Grounding~\cite{videogrounding}, Dense Video Captioning~\cite{densevideocaption}, and Video Highlight Detection~\cite{videogrounding}), and 3) object-level tasks (\textit{e.g.}, Single Object Tracking (SOT)~\cite{siamfc,siamrpn}, Multi-Object Tracking (MOT)~\cite{sort,deepsort}, and Video Object Segmentation (VOS)~\cite{davis2016,youtubevos}).
Video-level tasks are primarily focused on capturing global information within a video. Previous studies~\cite{videochatgpt,videollama,videollava,videochat,valley} have demonstrated the effectiveness of adapter architectures that employ fusion operations on the temporal axis, extracting features from all video frames, reducing the number of visual tokens at the expense of the capability for individual frame analysis. 
In contrast, frame-level tasks necessitate the model to differentiate and analyze each frame independently, emphasizing temporal awareness. Recent research~\cite{moviechat,timechat,vtimellm} has explored various tasks, including Video Grounding, Dense Video Captioning, and Video Highlight Detection. These endeavors have focused on leveraging temporal-aware token compression models to extend the context frames and capture longer-term temporal dependencies.
A step forward, object-level tasks require the model to differentiate and locate objects in each frame, ensuring temporal consistency with higher granularity. This capability is crucial for accurate trajectory tracking across multiple frames despite interference like occlusion and motion blur. Such tasks demand a sophisticated architecture that effectively distinguishes and tracks object identities over time while minimizing the use of visual tokens to achieve a larger context window. Moreover, training models for object-level tasks faces challenges due to the limited availability of large-scale training data, making it a challenging task for MLLMs to handle.

In this paper, we introduce Elysium, an end-to-end trainable MLLM that is capable of handling both global-level and object-level tasks in the video field. To address the issue of limited training data, we construct a large-scale dataset named \textbf{\dataset{}}, which includes abundant trajectories and object descriptions to support tasks like Single Object Tracking (SOT), RSOT, and Video-REG. \dataset{} is derived from the WebVid-10M dataset~\cite{webvid}, and we employ a carefully designed processing pipeline to reduce noise in the generated labels of each video.
To enable the MLLM to distinguish individual frames while reducing visual token use, we introduce a visual token compression network called \textbf{T-Selector}. This network offers a trade-off between performance and visual token consumption.
We leave the exploration of additional object-level tasks, such as Multi-Object Tracking, Video Object Detection, Video Object Segmentation, and Referring Video Object Segmentation (RVOS)~\cite{rvos,rvosactor,urvos} as future work. 
Our contributions can be summarized in three key aspects:

1). We construct \textbf{\dataset{}}, a million-scale dataset for object-level tasks in videos, supporting existing tasks like SOT and introducing benchmarks for RSOT and Video-REG.

2). We introduce \textbf{Elysium}, an end-to-end trainable MLLM, equipped with a carefully designed token compression network named \textbf{T-selector}. This approach extends the object-perception capabilities of MLLMs to encompass multiple frames, specifically videos.

3). Extensive experiments have shown the effectiveness of \model{} in downstream tasks such as Image Grounding, Video QA, SOT, RSOT, and Video-REG.
\section{Related Works}
\subsection{MLLMs}
In recent years, Multimodal Large Language Models have made remarkable advancements in a wide range of multi-modal tasks. Vision-language models~\cite{llava,blip2}, in particular, have achieved significant success in tasks such as image captioning and image question answering.
These models have demonstrated their ability to understand the visual content of images and videos and follow the instructions given by humans. This is achieved through the use of a lightweight adapter that connects the visual encoder with the language model. Typically, these models are pretrained on large-scale datasets and then finetuned on downstream task-aware datasets.
Furthermore, researchers have extended the capabilities of MLLMs~\cite{minigpt2,shikra,kosmos2,cogvlm,gpt4roi} to tackle object-level tasks in images, \textit{e.g.}, Object Detection and Image Grounding. These tasks require MLLMs to provide object coordinates as outputs, enabling accurate localization and identification of specific objects.
However, there has been a relative lack of research in the object-level tasks in videos, \textit{e.g.,} object tracking, except early exploration~\cite{pgvideollava} that employs multiple external expert models. Therefore, in this paper, we aim to extend the research to the domain of videos using an MLLM only. 
\subsection{Object-level Tasks in Videos}
In the domain of videos, there exist numerous object-level tasks, including Single Object Tracking, Multi-Object Tracking, Video Object Detection, Video Object Segmentation, and Referring Video Object Segmentation. Among these, Single Object Tracking stands as a fundamental task aimed at predicting the location of a specific object in consecutive frames by referencing its initial position in the first frame. Existing models~\cite{siamfc,siamfcpp,siamgat,siamrpnpp,mixformer} commonly employ two inputs: a template extracted from the region of interest in the first frame and a dynamic search area that adjusts based on the tracking result from the previous frame.
Typically, the search area takes the form of a larger square area centered around the predicted tracking box of the previous frame. This approach effectively narrows down the region where the object is likely to be located in subsequent frames.
To enhance tracking performance, some techniques involve periodically updating the template~\cite{mixformer,drol} to adapt to appearance changes, while others utilize a cosine window~\cite{siamfcpp,siamrpnpp} to constrain the heatmap and improve localization accuracy. However, these methods often require handcrafted parameter tuning to adapt to various scenes with different frame rates or fast-moving objects.
Moreover, Referring Video Object Segmentation~\cite{rvos} is a task that leverages language to locate and segment objects within videos, thereby combining visual and linguistic modalities to achieve precise object localization and segmentation. These object-level tasks play crucial roles in video analysis and understanding, enabling various applications. 
\section{Construct \dataset{} dataset}
In this section, we will give a clear definition about two tasks: RSOT and Video-REG. We will provide details on constructing the \dataset{} dataset and discuss its scale and the tasks it can support. Additionally, we will address the evaluation of model performance and the metrics used on the \dataset{} dataset.
\begin{figure}[t]
    \centering
    \includegraphics[width=\linewidth]{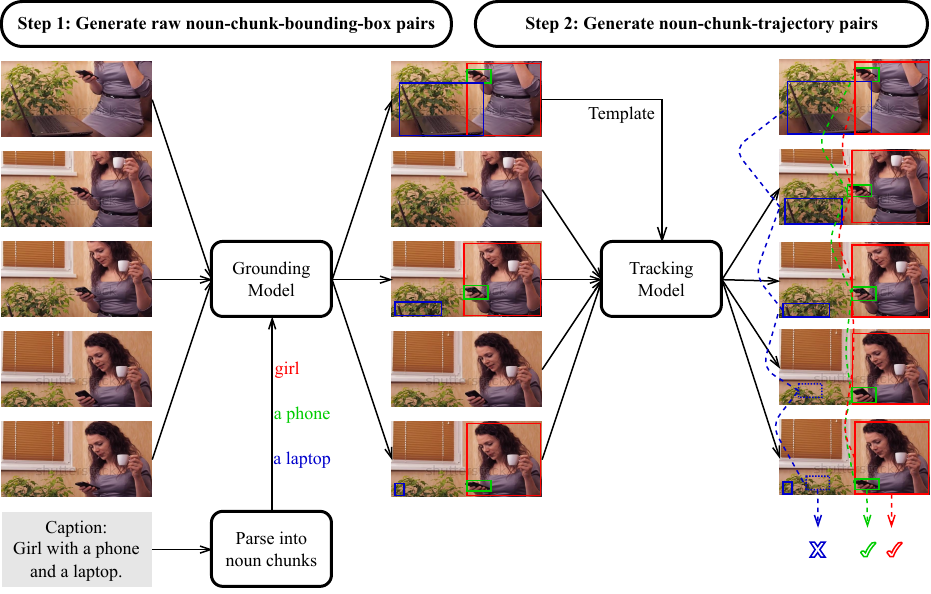}
    \caption{The pipeline to construct \dataset{} dataset.}
    \label{dataset}
\end{figure}
\begin{figure}[t]
    \centering
    \includegraphics[width=\linewidth]{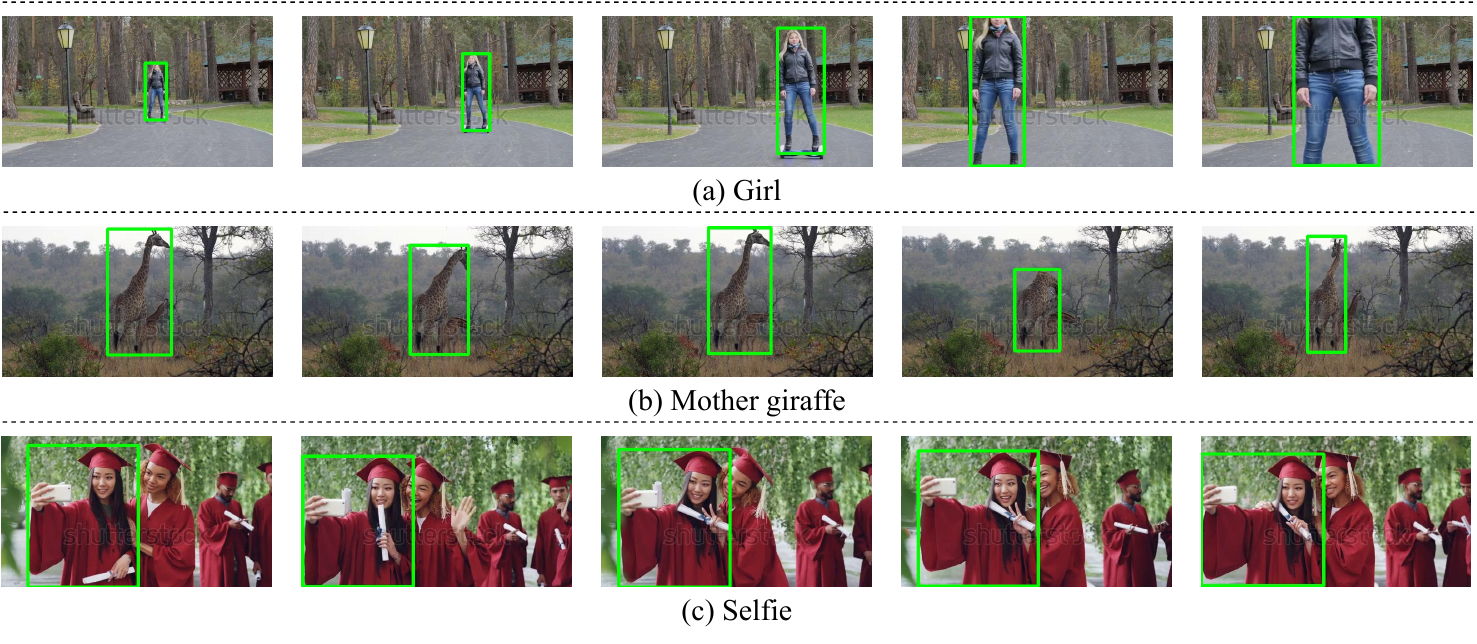}
    \caption{Samples from \dataset{} dataset.}
    \label{datasetsamples}
\end{figure}
\begin{table}[t]
    \centering
    \setlength{\tabcolsep}{6pt}
    \caption{Dataset comparisons on the aspects of 1) number of trajectories. 2) number of expressions. 3) total duration.}
    \begin{tabular}{l|ccc}
    \toprule
        \textbf{Datasets} & \textbf{\#Trajectories} & \textbf{\#Expressions} & \textbf{Duration}\\
    \midrule
        OTB15~\cite{otb15} & 51 & 0 & 16.4 minutes \\
        VOT14~\cite{vot14} & 25 & 0 & 5.7 minutes \\
        VOT16\cite{vot16} & 60 & 0 & 11.9 minutes \\
        VOT17\cite{vot17} & 60 & 0 & 11.9 minutes \\
        UAV20L~\cite{uav} & 20 & 0 & 32.6 minutes \\
        UAV123L~\cite{uav} & 91 & 0 & 1.1 hours \\
        GOT-10K~\cite{got10k} & 10K & 0 & 1.7 days \\
        LaSOT~\cite{lasot} & 1.4K & 1.4K & 1.4 days \\
        TrackingNet~\cite{trackingnet} & 30.6K & 0 & 5.6 days \\
    \midrule
        \textbf{\dataset{}} & \textbf{1.27M} & \textbf{1.27M} & \textbf{9.2 months} \\
    \bottomrule
    \end{tabular}
    
    \label{tab:dataset}
\end{table}
\subsection{Introduce RSOT and Video-REG Tasks}
\textbf{RSOT. }Motivated by Referring Video Object Segmentation, we define the task of RSOT as identifying and locating a specific object within an entire video by the given language expression only (no positional cue is used). This task offers a more flexible format for tracking compared to traditional Single Object Tracking methods and establishes a valuable connection between language and tracking.
\\
\textbf{Video-REG. }In addition to RSOT, we further extend the concept of Referring Expression Generation (REG) to the field of videos. We define this task as Video-REG, which involves predicting a description of an object given its coordinates in any frame of a video. 
Unlike conventional REG tasks, Video-REG demands the model to possess temporal awareness. This is because the appearance of the object in the current frame might be affected by occlusion or motion blur, but it can be identified in other frames provided. Thus, the model needs to consider the temporal context to generate precise descriptions in Video-REG.
\subsection{Dataset Construction}
We present \dataset{}, a million-scale object perception video dataset, designed to support SOT, RSOT, and Video-REG tasks. As depicted in Figure \ref{dataset}, the dataset construction pipeline consists of two essential steps: 1) Generating raw noun-chunk-bounding-box pairs and 2) Extending noun-chunk-bounding-box pairs to noun-chunk-trajectory pairs:
\\
\textbf{Step 1: Generating raw noun-chunk-bounding-box pairs.} Similar to prior research~\cite{kosmos2}, we employ spaCy~\cite{spacy} to parse the video captions in the WebVid-10M~\cite{webvid} dataset into noun chunks. To minimize noise, we remove chunks that contain virtual words (e.g., time, love, wind) or plural words (e.g., family, people, two dogs).
Next, we utilize a pretrained grounding model, Grounding DINO~\cite{groundingdino} to generate bounding boxes for each noun chunk in the first frame, middle frame, and last frame of a video. We retain noun-chunk-bounding-box pairs with predicted confidence scores higher than 0.6.
\\
\textbf{Step 2: Extending noun-chunk-bounding-box pairs to noun-chunk-trajectory pairs. } We utilize a pretrained Tracking Model (i.e., MixFormer~\cite{mixformer}) to generate trajectories given the bounding boxes in the first frame, thus extending raw noun-chunk-bounding-box pairs to raw noun-chunk-trajectory pairs. We retain noun-chunk-trajectory pairs that achieve a tracking confidence score higher than 0.8 consistently across all frames in a video. We also utilize the Kalman Filter~\cite{sort,deepsort} to discard trajectories with extreme drift. 
Subsequently, we calculate the Intersection over Union (IoU) score between the grounding box and the tracking box in the middle frame and the last frame. Cases where either of these scores is lower than 0.3, indicating potential tracker drift, are discarded to ensure the removal of inaccurate tracking instances.

In the end, 1.27 million noun-chunk-trajectory pairs in total are generated for \dataset{} dataset, samples are shown in Figure \ref{datasetsamples}. The whole process takes about 6 days on 24 A100-80G GPUs. As shown in Table \ref{tab:dataset}, it contains a significantly larger number of trajectories compared to existing tracking datasets, with each trajectory accompanied by an expression that refers to the corresponding object. The dataset is split into two subsets: around 1.27 million videos are used for training, while 500 videos are retained for evaluation. For the evaluation split, we manually check each case and ensured that the referring object in the initial frame of each video is unique.
%
\subsection{Evaluations}
For evaluation purposes on RSOT and SOT, we utilize the widely adopted One-Pass Evaluation (OPE) strategy as described in~\cite{sotmetrics}. We use commonly used metrics such as success and precision, which serve as the evaluation criteria for both traditional SOT tasks and the proposed RSOT task on the \dataset{} dataset. By employing these established metrics, we ensure a consistent and standardized evaluation process, enabling fair comparisons and accurate assessments of tracking performance across different tasks. For Video-REG tasks, we use Meteor~\cite{meteor} and CIDEr~\cite{cider} for evaluation following~\cite{kosmos2}.
\section{Elysium}
\begin{figure}[!t]
    \centering
    \includegraphics[width=\linewidth]{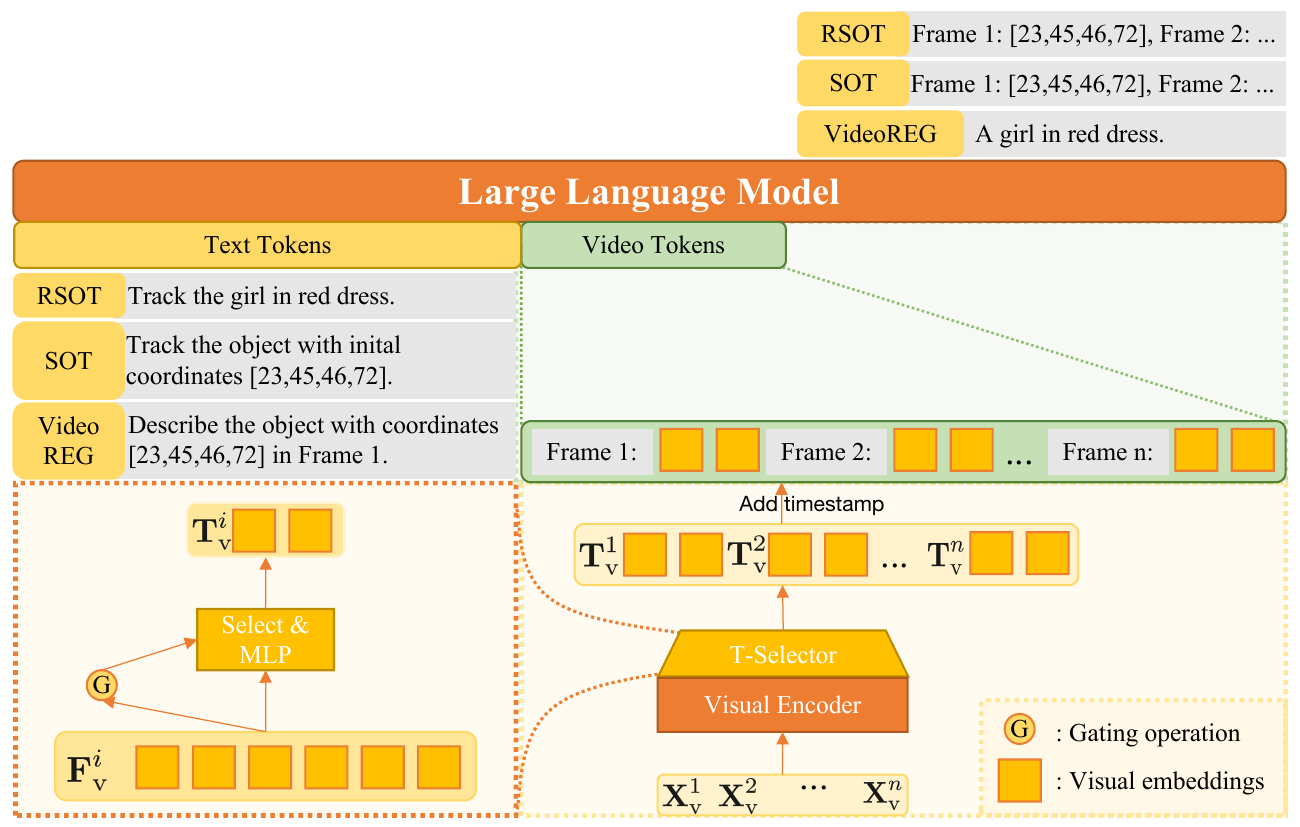}
    \caption{The architectures of Elysium. Elysium combines a visual encoder, an LLM, and a T-Selector to connect the visual encoder with the LLM.}
    \label{elysium}
\end{figure}
\subsection{Architecture}
\textbf{Overall architecture.} The primary goal of architectural designs in the context of MLLMs encompasses two key aspects. Firstly, it aims to enhance the MLLMs' ability to tackle object-level tasks in videos, such as object tracking. This entails addressing challenges related to incorporating temporal information and effectively handling inter-frame relationships within the MLLM framework. Secondly, architectural designs need to strike a balance between minimizing the number of tokens consumed by visual features and optimizing the overall performance of the MLLM on downstream tasks. Thus, we choose CLIP-ViT-L~\cite{clip} as our visual encoder, Vicuna~\cite{vicuna2023} as our LLM, and a specially designed token compressor (T-selector) to connect visual encoder and LLM.

As shown in Figure \ref{elysium}, for each frame $\mathbf{X}_{\mathrm{v}}^i$ in a video, we utilize a visual encoder to obtain its feature $\mathbf{F}_{\mathrm{v}}^i \in \mathbb{R}^{N \times C}$. Subsequently, a token-compressing model, denoted as T-Selector, compresses $\mathbf{F}_{\mathrm{v}}^i \in \mathbb{R}^{N \times C}$ into $\mathbf{T}_{\mathrm{v}}^i \in \mathbb{R}^{\alpha N \times D}$, where $\alpha \in (0,1]$ is a hyperparameter that indicates the compression ratio of the token count, and $D$ represents the hidden dimension of the Large Language Model (LLM). The process can be expressed as follows:
\begin{equation}
\begin{aligned}
\mathbf{F}_{\mathrm{v}}&=g\left(\mathbf{X}_{\mathrm{v}}\right), \mathbf{F}_{\mathrm{v}} \in \mathbb{R}^{N \times C}; \\
\mathbf{T}_{\mathrm{v}}&=h\left(\mathbf{F}_{\mathrm{v}}\right), \mathbf{T}_{\mathrm{v}} \in \mathbb{R}^{\alpha N \times D},
\end{aligned}
\end{equation}
where $g$ indicates the visual encoder (\textit{i.e.}, CLIP-ViT-L) and $h$ is the T-Selector.
\\
\textbf{T-Selector. } Given the hypothesis that videos contain redundant information~\cite{videomae,maest}, our research focuses on investigating effective methods to compress visual tokens. The goal is to enable the MLLM to handle a larger number of frames without suffering a significant drop in performance. We have explored various visual token compression techniques, including commonly used architectures such as one-layer cross-attention and concatenation. Notably, we have observed that fusion operations on the spatial dimension result in a drastic performance decline. To mitigate this issue, we propose a novel approach called T-selector. The T-selector aims to strike a balance between the visual token count and performance. It comprises two parts: a gating operation conducted by an MLP layer and a softmax layer, responsible for determining which tokens to select, and an MLP layer that transforms the hidden dimension to match that of the LLM. The gating process can be written as: 
\begin{equation}
\begin{gathered}
\mathbf{G}_{\mathrm{v}}=\operatorname{KeepTopK}(\operatorname{Softmax}(\operatorname{MLP}(\mathbf{F}_{\mathrm{v}})), k, \mathbf{F}_{\mathrm{v}}), \\
\mathbf{T}_{\mathrm{v}}=\operatorname{MLP}(\mathbf{G}_{\mathrm{v}}),
\end{gathered}
\end{equation}
\noindent where $k = \alpha N$, indicating the compression ratio of T-Selector. With a gating operation denoted as $\operatorname{KeepTopK}$, we select the tokens with scores among the top $k$, the selected tokens are represented as $\mathbf{G}_{\mathrm{v}}$. This selection strategy helps to reduce the number of tokens while maintaining the most relevant and informative ones according to their scores. Visual tokens after MLP layer are denoted as $\mathbf{T}_{\mathrm{v}}$.
\\
\textbf{Input \& output format. } As depicted in Figure \ref{elysium}, we have incorporated timestamps into each visual token $\mathbf{T}_{\mathrm{v}}^i$ to enable Elysium to differentiate between consecutive frames. To optimize token use, we represent a location simply using the coordinates of the top-left and bottom-right corners. Each value is in the range of $[0, 100)$, separated by a comma without extra space. For instance, a coordinate like ``[23,45,46,72]'' requires 13 tokens only in LLaMA~\cite{llama} tokenizer, which is much fewer compared to ``[0.231, 0.452, 0.458, 0.783]'' used by Shikra\cite{shikra} (28 tokens). Additionally, we have devised distinct prompts tailored to different types of questions and expanded the question set to include multiple queries for each task. This approach enhances Elysium's robustness when dealing with diverse question formats.
\subsection{Training Setups}
\textbf{Datasets Preparation.} Following previous works~\cite{kosmos2,minigpt2,shikra,qwenvl}, we use large-scale data to train our model. We classify the datasets we use into three classes: 1) Image data: LLaVA-558K~\cite{llava15}, GRIT-20m~\cite{kosmos2}, CogVLM-SFT-311K~\cite{cogvlm}, COCOVQA~\cite{cocovqa}, GQA~\cite{gqa}, Visual7W~\cite{visual7w}, Flickr~\cite{flickr30k}, VisualGenome~\cite{visualgenome}, RefCOCO~\cite{refcoco}, RefCOCO+~\cite{refcoco}, RefCOCOg~\cite{refcocog}; 2) Data for video-level tasks: VideoChat~\cite{videochat}; 3) Video data for object-level tasks: \dataset{}.

To achieve comprehensive training for an MLLM, we employ progressive training strategies. The overall training process involves two distinct stages: pretraining and finetuning stage.
\\
\textbf{Stage 1: Pretraining on large-scale image data. } For pretraining, we exclusively utilize image data due to their smaller token count, resulting in faster training speeds. To ensure stable training, we adopt a two-step process. Initially, we employ LLaVA-558K to exclusively train the T-Selector. During this phase, the ViT and LLM components remain frozen. The learning rate is set to $2\times10^{-3}$, the batch size is 32, and the training is performed on 8 GPUs. This step aims to initialize the T-Selector's parameters, enhancing the stability of subsequent training procedures.
Following the T-Selector training, we proceed to train Elysium end-to-end, unfreezing all parameters. This training involves a mixture of image data and is conducted for 30,000 steps on 32 GPUs. The learning rate is set to $5 \times 10^{-5}$, and the batch size is 16.
\\
\textbf{Stage 2: Finetuning on high-quality data. } After the pretraining phase, we proceed to train Elysium using high-quality data to enhance its performance in object-level tasks and downstream applications like image grounding. This training involves a mixture of datasets comprising high-quality image data and video data, \textit{i.e.}, VideoChat, \dataset{}, and all the image datasets mentioned above except GRIT-20M and LLaVA-558K. To optimize the training process, we conduct training for 22,000 steps on 32 GPU cards, utilizing a learning rate of $5 \times 10^{-5}$ and a batch size of 16. 
For the video dataset, we adopt a random sampling approach, randomly selecting 2 to 8 frames per video with a random interval ranging from 1 to 60. This selection process helps simulate varying frame rates and motion speeds. This configuration is used for the first 20,000 steps. In the subsequent 2,000 steps, we extend the frame count to 32 frames per video, utilizing a batch size of 1.
By employing this training methodology, we aim to optimize the performance of Elysium in object-level tasks and ensure its adaptability to different video scenarios with varying frame rates and motion speeds.

During the training process, we resize the images to a size of $336 \times 336$ without padding. Data augmentations are not applied. To optimize training, we utilize a cosine learning rate schedule. Additionally, we employ deepspeed~\cite{deepspeed} to leverage larger batch sizes, enhancing computational efficiency. All experiments are conducted on A100-80G GPUs.
\subsection{Evaluation Setups}
For video-level tasks like VideoQA, we uniformly sample 16 frames from a video for inference. However, for object-level tasks such as RSOT and SOT, if the video has more than 32 frames, we divide it into smaller clips. Each clip consists of 8 frames, with an overlap of one frame in the preceding and subsequent clips. Thus, by utilizing the predicted location of the last frame in the previous clip, we can initialize the tracking process in the subsequent clip. This process aligns with the commonly used updating template trick in SOT. 
\section{Experiments}
\subsection{State-of-the-art Comparisons}
\begin{table}[t]
    \centering
    \setlength{\tabcolsep}{0.5pt}
    \caption{Results on Referring Expression Comprehension. ``\textit{Res.}'' indicates the resolution of the input image. ``\textit{T.}'' denotes the visual token count per image. ``*'' represents this method utilizes a much larger ViT~\cite{eva} than CLIP-ViT-L. }
    \begin{tabular}{lll|ccc|ccc|cc}
        \toprule
        \multicolumn{3}{c|}{\textbf{Model}} & \multicolumn{3}{c}{\textbf{RefCOCO}} & \multicolumn{3}{c}{\textbf{RefCOCO}+} & \multicolumn{2}{c}{\textbf{RefCOCOg}} \\
        \cline{1-11}
        Model        & \textit{Res.}     & \textit{T.}  & val   & test-A    & test-B    & val   & test-A    & test-B    & val   & test\\
        \midrule
        OFA-L~\cite{ofa}        & -      & -   & 79.96 & 83.67     & 76.39     & 68.29 & 76.00     & 61.75     & 67.57 & 67.58\\ 
        VisionLLM-H~\cite{visionllm}  & -    & -    & -     & 86.70     & -     & - & -     & 61.75     & - & -\\ 
        Shikra (7B)~\cite{shikra}   & 224      & 256    & 87.01 & 90.61     & 80.24     & 81.60 & 87.36     & 72.12     & 82.27 & 82.19\\   
        Shikra (13B)~\cite{shikra}  & 224      & 256    & 87.83 & 91.11     & 81.81     & \textbf{82.89} & 87.79     & 74.41     & 82.64 & 83.16\\ 
        MiniGPT-v2 (7B)$^*$~\cite{minigpt2}  & 448      & 256    & 88.69 & 91.65     & \textbf{85.33}     & 79.97 & 85.12     & 74.45     & \textbf{84.44} & 84.66\\ 
        Ferret (7B)~\cite{ferret}   & 336      & 608    & 87.49 &  91.35     &  82.45    & 80.78 & 87.38     & 73.14     & 83.93 & \textbf{84.76}\\
        GroundingGPT (7B)~\cite{groundinggpt}   & 336      & 576    & 88.02 &   91.55     &  82.47    & 81.61 & 87.18    & 73.18     & 81.67 & 81.99\\
        \midrule
        \textbf{Elysium} (\textit{ours}, 7B)  & 336      & 108    & \textbf{89.07} & \textbf{92.12}     & 84.95     & 82.86 & \textbf{88.93}     & \textbf{75.64}     & 82.92 & 83.62\\
        \bottomrule
    \end{tabular}
    
    \label{tab:refcoco}
\end{table}
\begin{table}[t]
    \centering
    \setlength{\tabcolsep}{1pt}
    \caption{Results on VideoQA. Our Elysium achieves state-of-the-art performance on the public datasets.}
    \begin{tabular}{l|cc|cc|cc|cc}
        \toprule
        \multirow{2}{*}{\textbf{Model}} & \multicolumn{2}{c|}{\textbf{MSVD-QA}} & \multicolumn{2}{c|}{\textbf{MSRVTT-QA}} & \multicolumn{2}{c|}{\textbf{TGIF-QA}} & \multicolumn{2}{c}{\textbf{ActivityNet-QA}}\\
        \cline{2-9}
                    & Acc.  & Score     & Acc.  & Score     & Acc.  & Score     &   Acc.    & Score \\
        \midrule
        FrozenBiLM~\cite{frozenbilm}  & 32.2      & -         & 16.8      & -         & 41.0      & -     &               24.7       & - \\
        VideoLLaMA~\cite{videollama}  & 51.6      & 2.5       & 29.6      & 1.8       & -      & -     &               12.4       & 1.1 \\
        LLaMA-Adapter~\cite{llamaadapter}  & 54.9      & 3.1       & 43.8      & 2.7       & -      & -     &               34.2       & 2.7 \\
        VideoChat~\cite{videochat}  & 56.3      & 2.8       & 45.0      & 2.5       & 34.4      & 2.3     &               26.5       & 2.2 \\
        VideoChatGPT~\cite{videochatgpt}  & 64.9      & 3.3       & 49.3      & 2.8       & 51.4      & 3.0     &               35.2       & 2.7 \\
        Valley-v3~\cite{valley} & 60.5      & 3.3       & 51.1      & 2.9       & -      & -     &               45.1       & 3.2 \\
        MovieChat~\cite{moviechat} & 75.2      & \textbf{3.8}       & 52.7      & 2.6       & -      & -     &               \textbf{45.7}       & \textbf{3.4}\\
        \midrule
        \textbf{Elysium} (\textit{ours}) & \textbf{75.8}  & 3.7 & \textbf{67.5} & \textbf{3.2} & \textbf{66.6} & \textbf{3.6} & 43.4 & 2.9 \\
        \bottomrule
    \end{tabular}
    
    \label{tab:videoqa}
\end{table}
\begin{table}[t]
    \centering
    \setlength{\tabcolsep}{6pt}
    \caption{Results on Single Object Tracking. Our Elysium conducted a \textbf{zero-shot} experiment on the tracking datasets without any finetuning. All other methods are finetuned on corresponding datasets.}
    \begin{tabular}{l|ccc|cc}
        \toprule
        \multirow{2}{*}{\textbf{Model}} & \multicolumn{3}{c|}{\textbf{LaSOT}} & \multicolumn{2}{c}{\textbf{UAV123}}\\
        \cline{2-6}
         & \textbf{AUC} & \text{\textbf{P}} & $\text{\textbf{P}}_{\text{Norm}}$ & \textbf{AUC} & \text{\textbf{P}} \\
        \midrule
        SiamFC~\cite{siamfc} & 33.6 &  42.0 & 33.9 & 48.5 & 69.3 \\
        SiamRPN~\cite{siamrpn} & - &  - & - & 52.7 & 74.8 \\
        MDNet~\cite{mdnet} & 39.7 &  46.0 & 37.3 & 52.8 & - \\
        SiamRPN++~\cite{siamrpnpp} & 49.6 &  56.9 & 49.1 & 61.0 & 80.3 \\
        DiMP~\cite{dimp}  & \textbf{56.9} &   \textbf{65.0} & \textbf{56.7} & \textbf{65.4 }& - \\
        MAML~\cite{maml}  & 52.3 &  - & - &  - & - \\
        SiamFC++~\cite{siamfcpp}  & 54.4 &   62.3 &  54.7 & - & - \\
        CGACD~\cite{cgacd}  & 51.8 & 62.6 & - & 63.3 & 83.3 \\
        SiamGAT~\cite{siamgat}  & 53.9 &  63.3 &  53.0 & 64.6 & \textbf{84.3} \\
        \midrule
        \textbf{Elysium} (\textit{ours, zero-shot}) & 56.1  & 61.0 & 50.1 & 56.6 & 79.2 \\
        \bottomrule
    \end{tabular}
    
    \label{tab:sot}
\end{table}
\begin{table}[t]
    \centering
    \setlength{\tabcolsep}{6pt}
    \caption{The results of the RSOT and SOT tasks on the evaluation split of \dataset{} dataset are presented. As a baseline method, we employ MiniGPT-v2 to perform REC tasks on each frame of the videos.}
    \begin{tabular}{ll|ccc}
        \toprule
        \textbf{Model} & \textbf{Task} & \textbf{AUC} & \textbf{\text{P}} & $\text{\textbf{P}}_{\text{Norm}}$\\
        \midrule
        MiniGPT-v2~\cite{minigpt2} & RSOT & 65.4 & 70.1 & 67.4 \\
        \midrule
        \textbf{Elysium} (\textit{ours})  & RSOT & 87.5 & 94.5 & 93.7 \\
        \textbf{Elysium} (\textit{ours}) & SOT & \textbf{88.7} & \textbf{94.6} & \textbf{93.8} \\
        \bottomrule
    \end{tabular}
    \label{tab:rsot}
\end{table}
\begin{table}[t]
    \centering
    \setlength{\tabcolsep}{6pt}
    \caption{The results of the Video-REG tasks on the evaluation split of \dataset{} dataset. As a baseline, we employ MiniGPT-v2 to perform REG tasks on each given frame.}
    \begin{tabular}{l|cc}
        \toprule
        \textbf{Model} & \textbf{Meteor} & \textbf{CIDEr} \\
        \midrule
        MiniGPT-v2~\cite{minigpt2} & 16.9 & 76.6 \\
        \midrule
        \textbf{Elysium} (\textit{ours}) &\textbf{46.7} & \textbf{115.3} \\
        \bottomrule
    \end{tabular}
    \label{tab:videoreg}
\end{table}
In this section, we assess the capabilities of our proposed model, Elysium. All the experiments are conducted using the checkpoint after the second stage of training without any further finetuning. Firstly, we evaluate Elysium's object-level perception ability in images by examining its performance on image grounding tasks. Secondly, we gauge its proficiency in handling video-level tasks by testing its performance on VideoQA datasets. Lastly, we evaluate Elysium's object-level perception ability in videos through tasks such as Single Object Tracking. By conducting evaluations across these different task types, we aim to comprehensively assess the effectiveness of Elysium in various domains and levels of perception.
\\
\textbf{Image Grounding. } Image Grounding is a task that necessitates the model to produce the corresponding coordinates based on a given language expression. We evaluate the performance of our model, Elysium, on commonly used datasets such as RefCOCO~\cite{refcoco}, RefCOCO+~\cite{refcoco}, and RefCOCOg~\cite{refcocog} datasets. The performance results of Elysium are presented in Table \ref{tab:refcoco}. Despite employing a visual token compression method and representing images with fewer tokens, Elysium achieves state-of-the-art performance. This outcome validates the effectiveness of our token compression network in maintaining high performance while reducing token use.
\\
\textbf{Zero-Shot VideoQA Evaluation. } Following previous protocols~\cite{videochat,videochatgpt,valley}, we evaluate Elysium's video question-answering ability on multiple publicly available datasets, \textit{i.e.,} MSVD-QA~\cite{chen2011msvd}, MSRVTT-QA~\cite{xu2017msrvttqa}, TGIF-QA~\cite{jang2017tgif}, and ActivityNet-QA~\cite{yu2019activityqa}. To ensure a fair evaluation, we employ a zero-shot approach and utilize GPT-assisted evaluation techniques~\cite{videochatgpt}. This evaluation process involves assessing the accuracy of the model's generated predictions and assigning a relative score on a scale of 1 to 5. We aim to comprehensively evaluate and compare the performance of Elysium across different video question-answering datasets. As depicted in Table \ref{tab:videoqa}, our Elysium achieves state-of-the-art performance.
\\
\textbf{Zero-Shot SOT Evaluation. } We perform zero-shot evaluations on various SOT datasets using our Elysium model. As illustrated in Figure \ref{tab:sot}, Elysium demonstrates comparable performance to baseline methods, even in a zero-shot setting. However, we observe that Elysium's performance is relatively less satisfactory when dealing with datasets containing small objects, such as UAV123. This might be attributed to the limited resolution of the visual encoder we use.
\\
\textbf{Evaluation on \dataset{}.} As shown in Table \ref{tab:rsot} and Table \ref{tab:videoreg}, we evaluate the performance of RVOT, VOT, and VideoREG on the proposed dataset. In the RSOT task, the baseline method involves directly using MiniGPT-v2 to predict the coordinates of every frame based on the given expression, and the coordinates of each frame in the whole video are used for calculating metrics. In the Video-REG task, we utilize MiniGPT-v2 to generate the expression based on the given coordinates in the frame. The superior performance of Elysium, compared to MiniGPT-v2, can be attributed to its ability to capture temporal awareness and coherence, which are learned from the training data in \dataset{}.
\subsection{Ablation Studies}
\begin{table}[t]
    \centering
    \setlength{\tabcolsep}{2pt}
    \caption{Ablation studies on the architecture of visual token compression network. $Concat.$ and $C.A.$ stands for concatenation and cross-attention operation, respectively. \textit{None} represents not using any compression model.}
    \begin{tabular}{lcc|ccc|ccc|cc|c}
        \toprule
        \multicolumn{3}{c|}{\textbf{Architectures}} & \multicolumn{3}{c}{\textbf{RefCOCO}} & \multicolumn{3}{c}{\textbf{RefCOCO}+} & \multicolumn{2}{c|}{\textbf{RefCOCOg}} & \multirow{2}{*}{\textbf{Avg.}} \\
        \cline{1-11}
        Model       & Res.  & $\alpha N$    & val   & test-A    & test-B    & val   & test-A        & test-B    & val   & test\\
        \midrule
        \textit{None}      & 224      & 256  & 80.67 & 88.47 & 73.11 & 74.17 & 83.41 & 63.55 & 74.84 & 76.26 & 76.81\\
        \textit{None}      & 336      & 576  & \textbf{86.19} & \textbf{91.2}7 & \textbf{79.92} & \textbf{78.17} & \textbf{85.46} & \textbf{69.28} & \textbf{80.02} & \textbf{81.28} & \textbf{81.45}\\
        \midrule
        $Concat.$ & 336   & 144        & 59.29 & 68.29     & 50.28     & 50.85  & 60.30   & 42.28     & 52.76 & 53.18 & 54.65\\
        $C.A.$ & 336   & 144        & 52.24 & 55.52     & 50.40     & 37.23  & 40.85    & 32.81     & 40.52 & 40.60 & 49.23\\
        T-Selector  & 336   & 1             & 51.57 & 54.01     & 50.35     & 37.31  & 40.28    & 35.25     & 39.77 & 41.05 & 43.70\\
        T-Selector  & 336   & 36            & 81.39 & 85.47     & 76.30     & 70.84  & 76.12    & 61.91     & 71.59 & 72.20 & 74.48\\
        T-Selector  & 336   & 72            & 82.93 & 85.82     & 78.85     & 72.86  & 77.06    & 65.26     & 73.65 & 73.91 & 76.29\\
        T-Selector  & 336   & 108           & 84.98 & 88.24     & 80.15     & 75.77  & 79.27    & 67.41     & 76.29 & 76.21 & 78.54\\
        T-Selector  & 336   & 144           & 84.77 & 88.42     & 79.56     & 74.80  & 80.61    & 66.98     & 76.08 & 76.36 & 78.45\\
        T-Selector  & 336   & 256           & \textbf{86.25} & \textbf{90.06}     & \textbf{81.04}     & 77.03  & 82.43    & \textbf{68.09}     & 77.61 & 78.21 & \textbf{80.09}\\
        T-Selector  & 336   & 288           & 86.18 & 89.29     & 80.37     & \textbf{77.10}  & \textbf{82.62}    & 67.55     & \textbf{77.70} & \textbf{78.36} & 79.90\\
        
        \bottomrule
    \end{tabular}
    
    \label{tab:compressor}
\end{table}
We conducted ablation studies to explore the impact of adapter architecture and the visual token count between the LLM and the visual encoder on object perception performance in images. The models are first pretrained on the LLaVA-558K~\cite{llava15} and then finetuned on the RefCOCO~\cite{refcoco}, RefCOCO+~\cite{refcoco}, and RefCOCOg~\cite{refcocog} datasets.

Table \ref{tab:compressor} presents the results of our experiments. We keep the visual token count per image constant, and observe that fusion operations such as cross-attention~\cite{qwenvl} and concatenation~\cite{minigpt2} do not achieve promising performances compared to our T-Selector structure. Furthermore, we find that when the visual token count is the same, using a ViT@224p combined with an MLP achieves worse performance compared to using a ViT@336p along with our T-Selector architecture.

Additionally, we investigate the influence of the compression ratio on the final performance. We observe a trend where the performance degrades as the compression ratio $\alpha$ decreases. Specifically, we notice a significant performance drop when the visual token count per image $\alpha N$ decreases from 108 to 72. Therefore, to achieve a trade-off between a larger context window and performance, we select the value of $\alpha N = 108$ as default setting for subsequent experiments.
\subsection{Visualizations}
\begin{figure}[ht]
    \centering
    \includegraphics[width=\linewidth]{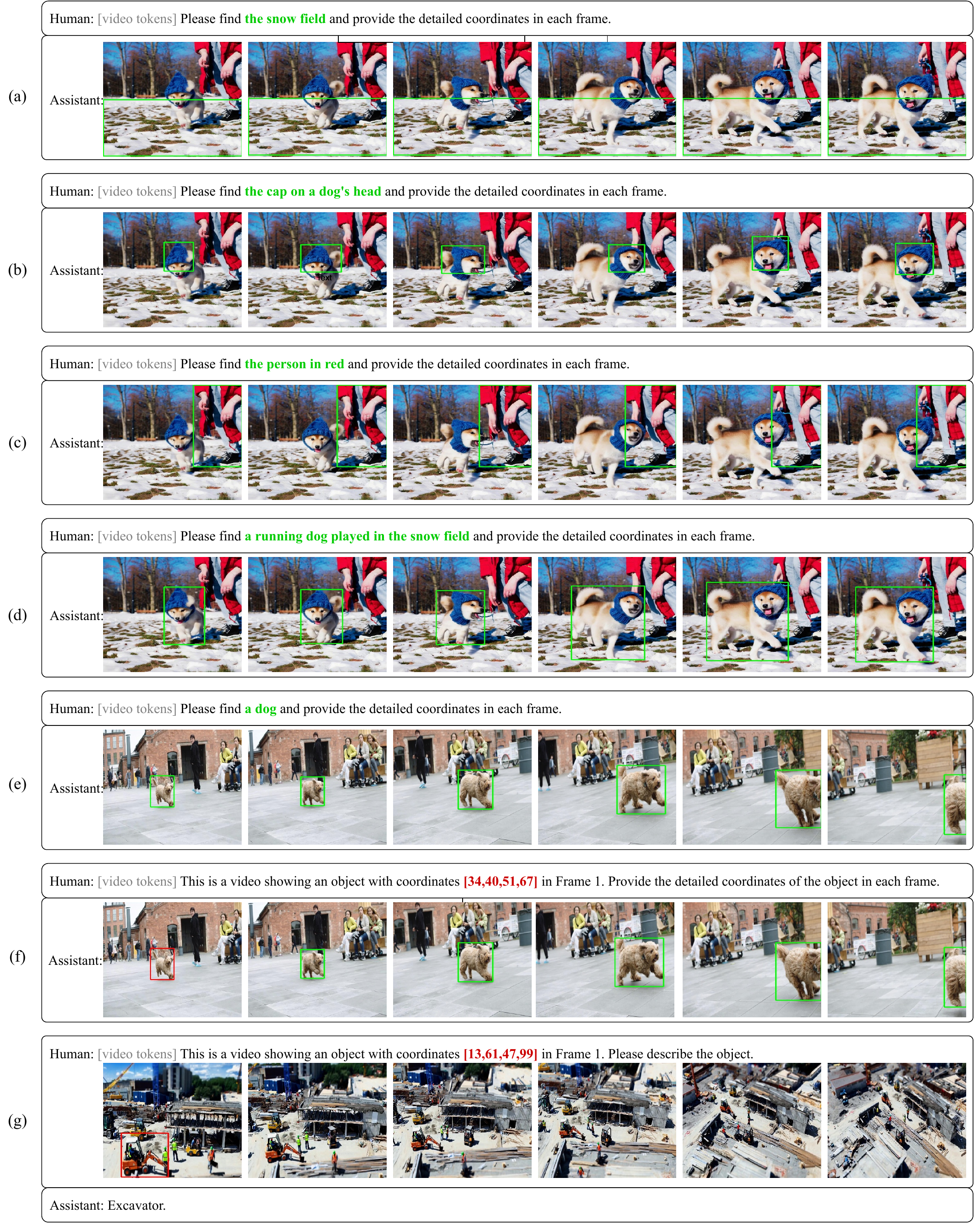}
    \caption{Visualizations of the Elysium. We showcased the performance of Elysium on several videos. We demonstrated the RSOT capability of Elysium on cases (a) to (e) while showcasing the SOT on case (f) and Video-REG on case (g). }
    \label{visualizations}
\end{figure}
As depicted in Figure \ref{visualizations}, we gather videos from Sora~\cite{sora} and a video webpage~\cite{pexels} to ensure that our Elysium model is not trained on such videos. In Figures \ref{visualizations}(a)-(e), we demonstrate the capability of our Elysium model to effectively handle various expressions of different lengths in the context of RSOT. The model successfully focuses on and generates the coordinates of the corresponding object based on different expressions as cues.
Furthermore, Figures \ref{visualizations}(e) and \ref{visualizations}(f) illustrate how Elysium can accurately track objects by utilizing both expressions and bounding boxes as cues.
Lastly, Figure \ref{visualizations}(g) shows that Elysium is capable of processing frame coordinates and generating appropriate expressions accordingly.
%
\section{Conclusions and Limitations}
\textbf{Conclusions.} In this paper, we present Elysium, an end-to-end trainable MLLM designed to leverage the full potential of MLLM in object-level perception tasks, encompassing both images and videos, paired with two tasks, RSOT and Video-REG, which bridge the gap between language and tracking in videos. To facilitate these object-level tasks, we construct a large-scale dataset called \dataset{}, providing support for tasks such as SOT, RSOT, and Video-REG. Furthermore, we propose a visual token compression network, T-Selector, to strike a balance between a large context window and overall performance.
Through extensive experiments, we demonstrate that MLLM exhibits remarkable object perception abilities in videos. The results validate the effectiveness and potential of our proposed approach in leveraging MLLM for object-level perception tasks.
\\
\textbf{Limitations.} During our explorations, we have observed that Elysium's performance in tracking-related tasks is less satisfactory when dealing with tiny objects. However, we believe that this limitation can be addressed through further exploration, particularly by incorporating higher-resolution input images. Additionally, it is worth noting that our exploration thus far has primarily focused on tracking-related tasks within the realm of videos. As part of future work, we acknowledge the need to delve into other tasks such as VOS and RVOS, which present promising avenues for further research.
\clearpage
%
%
\bibliographystyle{splncs04}
\bibliography{egbib}
\end{document}